%% file: main.tex
\documentclass[a4paper, 10pt, conference]{ieeeconf}

\usepackage{FG2021}

\FGfinalcopy

\IEEEoverridecommandlockouts

\usepackage{times}
\usepackage{epsfig}
\usepackage{graphicx}
\usepackage{amsmath}
\usepackage{amssymb}
\usepackage{capt-of}
\usepackage{etoolbox}
\usepackage{booktabs}
\usepackage{microtype}

\usepackage{eso-pic}
\usepackage{xspace}

\usepackage[accsupp]{axessibility}

\usepackage[pagebackref=true,breaklinks=true,colorlinks,bookmarks=false]{hyperref}
\usepackage[capitalize]{cleveref}

\input{sections/commands}
\input{sections/teaser}

\title{\LARGE \bf
Human Pose Manipulation and Novel View Synthesis using Differentiable Rendering
}

\author{\parbox{16cm}{\centering
    {\large Guillaume Rochette$^{1}$, Chris Russell$^{2}$, Richard Bowden$^{1}$}\\
    {\normalsize
    $^{1}$ University of Surrey, $^{2}$ AWS T\"ubingen}}
}

\begin{document}

\maketitle

\input{sections/abstract}
\input{sections/introduction}
\input{sections/related_work}
\input{sections/methodology}
\input{sections/experiments}
\input{sections/conclusion}
\input{sections/acknowledgements}

{\small
\bibliographystyle{ieee}
\bibliography{references}
}

\newpage
\onecolumn

\appendices
\crefalias{section}{appendix}
\crefalias{subsection}{appendix}
\input{sections/appendix}

\end{document}

%% file: sections/commands.tex
\makeatletter
\DeclareRobustCommand\onedot{\futurelet\@let@token\@onedot}
\def\@onedot{\ifx\@let@token.\else.\null\fi\xspace}
\def\eg{\emph{e.g}\onedot}

\def\etal{\emph{et al}\onedot}
\makeatother

\newcommand*{\sbf}[3]{{#1}^\intercal {#2}^{-1} {#3}}
\newcommand*{\argmin}{\operatornamewithlimits{argmin}}
\newcommand*{\argmax}{\operatornamewithlimits{argmax}}

\newcommand*{\erfc}{\operatornamewithlimits{erfc}}

%% file: sections/teaser.tex
\input{figures/teaser}

%% file: figures/teaser.tex
\let\oldtwocolumn\twocolumn
\renewcommand*\twocolumn[1][]{%
    \oldtwocolumn[{#1}{
    \centering
    \includegraphics[width=\linewidth]{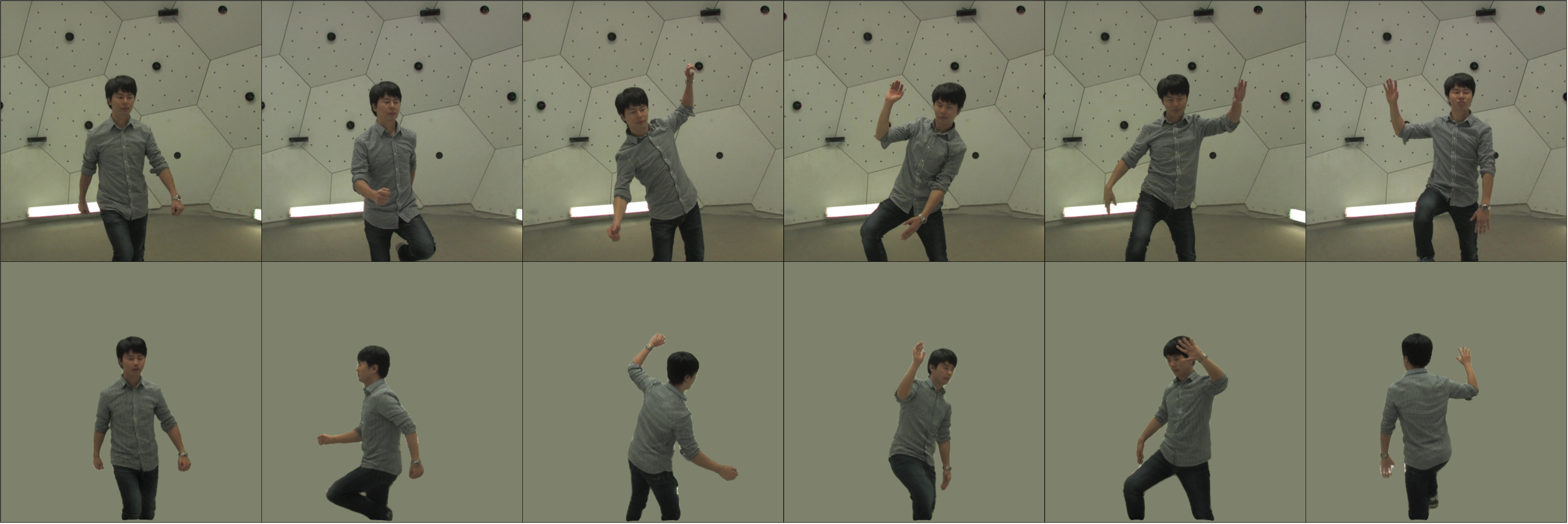}
    \captionof{figure}{From a previously unseen input image \textit{(top)}, we estimate the pose and appearance of the subject to render a novel view of the scene \textit{(bottom)}.}
    \label{fig:Teaser}
    }]
}

%% file: sections/abstract.tex
\begin{abstract}
    We present a new approach for synthesizing novel views of people in new poses.
    Our novel differentiable renderer enables the synthesis of highly realistic images from any viewpoint. Rather than operating over mesh-based structures, our renderer makes use of diffuse Gaussian primitives that directly represent the underlying skeletal structure of a human. Rendering these primitives gives results in a high-dimensional latent image, which is then transformed into an RGB image by a decoder network.
    The formulation gives rise to a fully differentiable framework that can be trained end-to-end.
    
    We demonstrate the effectiveness of our approach to image reconstruction on both the Human3.6M and Panoptic Studio datasets. We show how our approach can be used for motion transfer between individuals; novel view synthesis of individuals captured from just a single camera; to synthesize individuals from any virtual viewpoint; and to re-render people in novel poses.
    Code and video results are available at \url{https://github.com/GuillaumeRochette/HumanViewSynthesis}.
\end{abstract}

%% file: sections/introduction.tex
\section{Introduction}
\label{sec:Introduction}

We present a new three-step approach for novel view synthesis and motion transfer (see \cref{fig:Teaser}).
We first infer the pose and appearance information of a subject from an image.
The pose is then transferred to a novel view along with the appearance, from which we render diffuse primitives, using a realistic camera model, onto a high-dimensional image of the foreground of the scene.
These diffuse Gaussian primitives are semantically meaningful, simplify optimization and explicitly disentangle pose and appearance.
Finally, we use an encoder-decoder architecture to generate novel realistic images.
Leveraging  multi-view data, we train this framework end-to-end, optimizing both image reconstruction quality and pose estimation.
While our approach is generic and can be applied to many tasks, we focus on human pose estimation and novel view synthesis across large view changes.

\input{figures/overview}

Novel view synthesis is a fundamentally ill-posed problem, that we decompose into two parts.
The first involves solving an inverse graphics problem, requiring a deep understanding of the scene, while the second relies on image synthesis to generate realistic images using this understanding of the scene.
Differentiable rendering is an exciting area of research and offers a unified solution to these two problems. It allows the fusion of expressive graphical models, that capture the underlying physical logic of systems, with learning from gradient-based optimization.
Current approaches to differentiable rendering try to directly reason about the world by aligning a well-behaved and smooth approximate model with an underlying non-smooth, and highly nonconvex image.
While such approaches guarantee that gradients exist and that a local minimum can be found via continuous optimization, this is a catch-22 situation -- the more detailed the model, the better it can be aligned to represent the image; however, the more non-convex the problem becomes, the harder it is to obtain correct alignment.

Disentangling shape and appearance of an image is fundamental for general 3D understanding and lies at the heart of our idea.
If we focus on synthesizing novel views of humans, the localization of human joints in the three-dimensional space can be seen as a first step towards human body shape estimation.
If we simultaneously estimate the appearance of these body parts, then once we have extracted pose and appearance, we can transfer the information to a novel view and synthesize the output image.

We present a novel rendering primitive that allows us to take full advantage of recent progress in 3D human pose estimation and encoder-decoder networks to gain a higher degree of control over actor rendering, allowing us to synthesize known individuals from arbitrary poses and views. The key insight that allows this to take place is the use of diffuse Gaussian primitives to move from sparse 3D joint locations into a dense 2D feature image via a novel density renderer. An encoder-decoder network then maps from the feature space into RGB images (see \cref{fig:Overview}). By exploiting the simplicity of the underlying representation, we can generate novel views while `puppeting' the actors, re-rendering them in novel poses, as shown in \cref{fig:MotionTransfer}.      

\Cref{sec:RelatedWork} gives an overview of the literature related to human pose estimation and novel view synthesis.
\Cref{sec:Methodology} presents our approach for estimating the pose and synthesizing novel views of humans and our novel differentiable rendering.
\Cref{sec:Experiments} contains experiments performed on the Panoptic Studio and Human3.6M datasets, evaluating pose estimation accuracy and image reconstruction quality, as well as motion transfer between humans.

%% file: figures/overview.tex
\begin{figure*}[tbp]
    \centering
    \includegraphics[width=\linewidth]{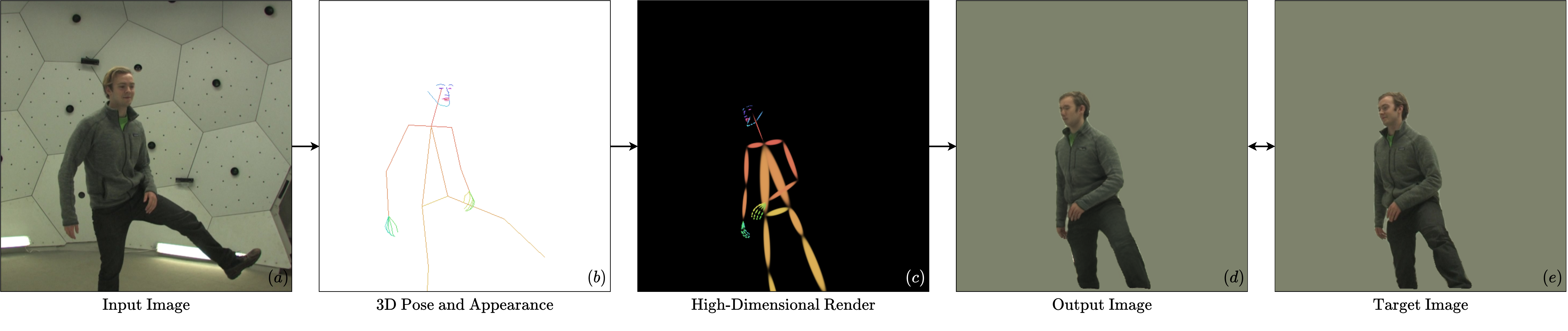}
    \caption{From an input image $I_{1}^{*}$, we infer the 3D joint locations $P_{1}$ and estimate appearance vectors $a_{1}$ for each limb (here, represented using an RGB palette). We transfer the pose from the input view to the novel view and render the primitives along with their appearances onto the high-dimensional latent image $J_{2}$. From the latent image, we synthesize the output image $I_{2}$, which should match the target image $I_{2}^{*}$.}
    \label{fig:Overview}
\end{figure*}

%% file: sections/related_work.tex
\section{Related Work}
\label{sec:RelatedWork}

\subsubsection{Human Pose Estimation}
Locating body parts in an image is inherently hard due to variability of
human pose, and ambiguities arising from occlusion, motion blur, and lighting.
Additionally, estimating depth is a
challenging task that is difficult even for humans due to perspective ambiguities.

Approaches to the 2D human pose estimation problem leverage recent advances in deep learning and large amounts of labeled data~\cite{lin2014microsoft}.
Convolutional Pose Machines~\cite{ramakrishna2014pose,wei2016convolutional} iteratively refine joint location predictions using the previous inference results.
Cao \etal~\cite{cao2017realtime} improved on this using part affinity fields and enabling multiperson scene parsing.

While 2D human pose estimation is robust to in-the-wild conditions, it is not the case for most 3D approaches, mainly due to the limitations of available data.
Large datasets use either Mo-Cap data, such as HumanEva~\cite{sigal2010humaneva} or Human3.6M~\cite{ionescu2013human3}, or rely on a high number of cameras for 3D reconstruction, such as Panoptic Studio~\cite{joo2017panoptic}.
These datasets are captured in tightly controlled environments and lack diversity, limiting the generalization of models.
Some approaches use multiobjective learning to overcome the lack of variability in the data, by either solving jointly for the 2D and 3D pose estimation tasks~\cite{li20143d}, or fusing 2D detection maps with 3D image cues~\cite{tekin2017learning}.
Recent trends revolve around learning 3D human pose with less constrained sources of data.
Inspired by Pose Machines~\cite{ramakrishna2012reconstructing,wei2016convolutional}, Tome \etal~\cite{tome2017lifting} trained a network with only 2D poses, by iteratively predicting 2D landmarks, lifting to 3D, and fusing 2D and 3D cues.
Martinez \etal~\cite{martinez2017simple} presented a simple 2D-to-3D residual densely-connected model that outperformed complex baselines relying on image data.
Drover \etal~\cite{drover2018can} proposed an adversarial framework, based on~\cite{martinez2017simple}, randomly reprojecting the predicted 3D pose to 2D with a discriminator judging the realism of the pose.
Chen \etal~\cite{chen2019unsupervised} refined~\cite{drover2018can} by adding cycle-consistency constraints.

\subsubsection{Image Synthesis}
Generative adversarial networks~\cite{goodfellow2014generative} are excellent at synthesising high-quality images, but typically offer little control over the generative process.
Progressively growing adversarial networks~\cite{karras2017progressive} was demonstrated to stabilize training and produce high-resolution photo-realistic faces.
Style modulation~\cite{karras2019style,karras2020analyzing} enabled the synthesis of fine-grained details, as well as a higher variability in the generated faces.
They can be conditioned to generate images from segmentation masks or sketches~\cite{isola2017image,wang2018high}.
Ma \etal~\cite{ma2017pose} proposed an adversarial framework allowing synthesis of humans in arbitrary poses, by conditioning image generation on an existing image and a 2D pose.
Chan \etal~\cite{chan2019everybody}, built on~\cite{wang2018high}, and proposed combining the motion of one human with the appearance of another, while preserving temporal consistency. However, it is limited between two fixed and similar viewpoints, due to the absence of 3D reasoning.

\subsubsection{Novel View Synthesis}
Novel view synthesis is the task of generating an image of a scene from a previously unseen perspective.
Most approaches rely on an encoder-decoder architecture, where the encoder solves an image understanding problem via some latent representation, which is subsequently used by the decoder for image synthesis.
Tatarchenko \etal~\cite{tatarchenko2016multi} presented an encoder-decoder using an image and a viewpoint as input to synthesize a novel image with depth information.
Park \etal~\cite{park2017transformation} refined it by adding a second stage hallucinating missing details.
Rather than synthesizing a novel view of images, Zhou \etal~\cite{zhou2016view} learnt the displacement of pixels between views.
Inspired by Grant \etal~\cite{grant2016deep}, Sitzmann \etal~\cite{sitzmann2019deepvoxels} used voxels to represent the 3D structure of objects and to deal with occlusion.
However, these approaches have difficulties dealing with large view changes, due to the unstructured underlying latent representation.

To account for larger view changes, Worrall \etal~\cite{worrall2017interpretable} structured their latent space to be directly parameterized by azimuth and elevation parameters.
Rhodin \etal~\cite{rhodin2018unsupervised} proposed a framework that disentangles pose, as a point cloud, and appearance, as a vector, for novel view synthesis and is later reused for 3D human pose estimation.
By explicitly structuring the latent space to handle geometric transformations, such models are able to handle larger view changes.
However, these approaches learn mappings to project 3D structures onto images, which may be undesirable as the mapping is specific to the nature of the rendered object.

\subsubsection{Differentiable Rendering}
Classical rendering pipelines, such as mesh-based renderers, are widely used in graphics.
They are good candidates for projecting information onto an image, however, they suffer from several major limitations.
Firstly, some operations such as rasterization are discrete and therefore not naturally differentiable.
As a substitute, one can use hand-crafted functions to approximate gradients~\cite{kato2018neural,loper2014opendr}.
However, surrogate gradients can add instability to the optimization process.
Liu \etal~\cite{liu2019soft} proposed a natively differentiable formulation by defining a `soft-rasterization' step.
Another problem with classical approaches is that representing objects as polygonal meshes for rendering purposes creates implicit constraints on the inverse graphics problem, as it requires the movement of vertices to preserve local neighbourhoods and importantly preserve nonlocal constraints of what is the \textit{inside} and \textit{outside} of the mesh.
This is challenging to optimize and requires strong regularization.
Shysheya \etal~\cite{shysheya2019textured} proposed learning texture maps for each body part to render novel views of humans, using 3D skeletal information as input.

\subsubsection{Neural Rendering}
Embedding a scene implicitly as a Neural Radiance Function (NeRF) is an emerging and promising approach.
Mildenhall \etal~\cite{mildenhall2020nerf} showed that it was possible to optimize an internal volumetric function using a set of input views of the scene.
Using sinusoidal activation functions, Sitzmann \etal~\cite{sitzmann2020implicit} enabled such networks to learn more complex spatial and temporal signals and their derivatives.
By combining neural radiance fields and the SMPL articulated body model, Su \etal~\cite{su2021nerf} proposed a model capable of generating human representations in unseen poses and views.
A range of dynamic NeRF variants have sprung up recently \cite{park2020deformable,pumarola2021d,tretschk2020non}, that simultaneously estimate deformation fields alongside the neural radiance function.
These approaches are shape agnostic, each model embeds the appearance of a single individual, and they are designed to run on short sequences and render novel views from a camera position similar to those previously captured.
In contrast, we explicitly train the model to learn the appearance of multiple individuals (29), allowing rendering from arbitrary views of a single frame captured separately, or for motion transfer from one individual to another.

%% file: sections/methodology.tex
\section{Methodology}
\label{sec:Methodology}
We jointly learn pose estimation as part of the view synthesis process, using our Gaussian-based renderer, as shown in \cref{fig:Overview}.
From the input image $I_{1}^{*}$, our model encodes two modalities, the three-dimensional pose $P_{1}$ relative to input camera, and the appearance $a_{1}$ of the subject.
The pose $P_{1}$ is transferred to a new viewpoint using camera extrinsic parameters $(R_{1 \rightarrow 2}, t_{1 \rightarrow 2})$.
From the pose $P_{2}$, seen from a novel orientation, we derive the location $\mu_{2}$ and shape $\Sigma_{2}$ of the primitives, which are used, along with their appearance $a_{1}$ and the intrinsic parameters and distortion coefficients of the second camera $(K_{2}, D_{2})$, for the rendering of the subject in a high-dimensional image $J_{2}$.
This feature image $J_{2}$ is enhanced in an image translation module to form the output image $I_{2}$, which closely resembles the target image $I_{2}^{*}$.

\subsection{Extracting Human Pose and Appearance from Images}
\label{sec:MethodologyFeatureExtraction}
We model the human skeleton as a graph $G = (P, E)$ with $N$ vertices and $M$ edges, where $P \in \mathbb{R}^{N \times 3}$ denotes the joint locations and $E = \{(i, j) | i, j \in [1..N]\}$.

\subsubsection{Inferring the Pose}
We use OpenPose~\cite{cao2017realtime}, an off-the-shelf detector, to infer the 2D pose from the input image $I_{1}^{*} \in \mathbb{R}^{H_{I} \times W_{I} \times 3}$,
\begin{align}
    I_{1}^{*}
    \rightarrow
    p_{1}^{*}, c_{1}^{*}
\end{align}
Here $p_{1}^{*} \in \mathbb{R}^{N \times 2}$ refers to the 2D joints locations and $c_{1}^{*} \in \mathbb{R}^{N}$ to their respective confidence values. \\
We use a simple fully-connected network~\cite{martinez2017simple} to infer the 3D pose from the 2D pose,
\begin{align}
    p_{1}^{*}, c_{1}^{*} \rightarrow \bar{P}_{1}
\end{align}
Where $\bar{P}_{1} \in \mathbb{R}^{(N - 1) \times 3}$ refers to the 3D pose relative to the root joint. \\
We compute the root joint location $P_{1, \text{root}}$ in camera coordinates by finding the optimal depth which minimize the error between the re-projected 3D pose $\bar{P}_{1}$ and the 2D pose $p_{1}^{*}$ (see \cref{appendix:Depth}).
We obtain the pose in camera coordinates $P_{1} \in \mathbb{R}^{N \times 3}$,
\begin{align}
    P_{1} = [P_{1, \text{root}} | P_{1, \text{root}} + \bar{P}_{1}]
\end{align}

\subsubsection{Estimating the Appearance}
From the input image $I_{1}^{*}$, we use a ResNet-50~\cite{he2016deep} to infer high-dimensional appearance vectors $a_{1}$ used to produce the primitives for rendering,
\begin{align}
    I_{1}^{*}
    \rightarrow
    a_{1}
\end{align}
Where $a_{1} \in \mathbb{R}^{M \times A}$ describe the appearances of each of the edges in the human skeleton as seen from the input view.
These high-dimensional vectors allow the transfer of the subject's appearance from one view to the other without considering pose configuration of the subject, and disentangling pose and appearance.

\subsubsection{Changing the Viewpoint}
To create an image of the world as seen from another viewpoint, we transfer the pose using the extrinsic parameters $(R_{1 \rightarrow 2}, t_{1 \rightarrow 2})$,
\begin{align}
    P_{2} = R_{1 \rightarrow 2} \times P_{1} + t_{1 \rightarrow 2}
\end{align}

\subsection{Differentiable Rendering of Diffuse Gaussian Primitives}
\label{sec:MethodologyRendering}
We render a simplified skeletal structure of diffuse primitives, directly obtained from the 3D pose.
The intuition underlying our new renderer is straightforward.
Each primitive can be understood as an anisotropic Gaussian defined by its location $\mu$ and shape $\Sigma$, and the rendering operation is the process of integrating along each ray.
Occlusions are handled by a smooth aggregation step.
Our renderer is differentiable with respect to each input parameter, as the rendering function is itself a composition of differentiable functions.

\subsubsection{From Pose to Primitives}
From the subject's pose, we compute the location and shape of the primitives,
\begin{align}
    P_{2}\rightarrow \mu_{2}, \Sigma_{2}
\end{align}
Where $\mu_{2} \in \mathbb{R}^{M \times 3}$ refers to the locations, while $\Sigma_{2} \in \mathbb{R}^{M \times 3 \times 3}$ refers to the shapes.
For clarity, we drop the subscripts indicating the viewpoint.

For each edge $(i, j)$, we create a primitive at the midpoint between two joints, with an anisotropic shape aligned with the vector between the two joints,
\begin{align}
    \mu_{i j} &= \frac{P_{i} + P_{j}}{2} \\
    \Sigma_{i j} &= R_{i j} \times \Lambda_{i j} \times R_{i j}^{\intercal}
\end{align}
Where,
\begin{align}
    R_{i j} &= f_{R}(P_{j} - P_{i}, e) \\
    \Lambda_{i j} &= 
    \text{diag}({||P_{j} - P_{i}||}_2, w_{i j}, w_{i j})
\end{align}
Here, $f_{R}$ calculates the rotation between two non-zero vectors (see \cref{appendix:Rotation}) and $w_{i j}$ loosely represents the width of the limb.

\subsubsection{Rendering the Primitives}
Modelling the scene as a collection of diffuse primitives, we render a high-dimensional latent image $J_{2}$,
\begin{align}
    J_{2} &= \mathcal{R}_{\alpha, \beta}(\mu_{2}, \Sigma_{2}, a_{1}, b, K_{2}, D_{2}) \in \mathbb{R}^{H_{R} \times W_{R} \times A}
\end{align}
Where
$\mathcal{R}$ is the rendering function;
$\alpha > 0$ is a coefficient scaling the magnitude of the shapes of the primitives;
$\beta > 1$ is a background blending coefficient;
$b \in \mathbb{R}^{A}$ describes the appearance of the background;
and $K_{2} \in \mathbb{R}^{3 \times 3}$ and $D_{2} \in \mathbb{R}^{K}$ refers to the intrinsic parameters and distortion coefficients.

To simplify notation, we drop the subscripts indicating the viewpoint, and retain the following letters for subscripts: $i \in [1..H_{R}]$ refers to the height of the image, $j \in [1..W_{R}]$ refers to the width of the image, and $k \in [1..M]$ refers to each of the $M$ primitives.

We define the rays $r_{i j}$ as unit vectors originating from the pinhole, distorted by the lens, and passing through every pixel of the image,
\begin{align}
    r_{i j} &= \frac{u(K^{-1} \times p_{i j}, D)}{{||u(K^{-1} \times p_{i j}, D)||}_{2}}
\end{align}
Where $u$ is a fast fixed-point iterative method~\cite{opencv_library} finding an approximate solution to undistorting the rays, and $p_{i j} = \begin{pmatrix} j & i & 1 \end{pmatrix}^{\intercal}$ is a pixel on the image plane.

Let $F_{i j k}$ be the integral of a single primitive $(\mu_{k}, \Sigma_{k})$ along the ray $r_{i j}$,
\begin{align}
    F_{i j k} &= \int_{0}^{+\infty}{e^{-\Delta^{2}(z \cdot r_{i j}, \mu_{k}, \alpha \cdot \Sigma_{k})} dz}
\end{align}
See \cref{appendix:Rendering_Integral} for the analytical solution.

For each ray $r_{i j}$, we define a smooth rasterisation coefficient $\lambda_{i j k}$ for each primitive $(\mu_{k}, \Sigma_{k})$.
In a nutshell, this coefficient smoothly favours one primitive, and discounts the others, based on their proximity to the ray $r_{i j}$.
See \cref{appendix:Smooth_Rasterization} for details.

The background is treated as the $(M + 1)^{\text{th}}$ primitive, with unique properties.
Its density $F_{i j M + 1}$ and smooth rasterisation coefficient $\lambda_{i j M + 1}$ are detailed in \cref{appendix:Background_Primitive}.

We derive the weights $\omega_{i j k}$ quantifying the influence of each primitive $(\mu_{k}, \Sigma_{k})$ (including the background) onto each ray $r_{i j}$, such that $\forall k \in [1..{M + 1}]$,
\begin{align}
    \omega_{i j k} &= \frac{\lambda_{i j k} \cdot F_{i j k}}{\sum\limits_{l=1}^{M + 1}{\lambda_{i j l} \cdot F_{i j l}}}
\end{align}
Finally, we render the image by combining the weights with their respective appearance,
\begin{align}
    J_{i j} &= \sum\limits_{k=1}^{M + 1}{\omega_{i j k} \cdot a_{k}}
\end{align}

\input{figures/infer_real}

\subsection{Synthesizing the Output Image}
\label{sec:MethodologyImageSynthesis}

The intermediate rendered image is feature-based and not photorealistic due to a small number of primitives.
While we could render any image with a sufficient number of primitives, increasing the number not only increases realism, but also the computational cost and the difficulty in optimising the overall problem.
Instead, we render a simplified skeletal structure $(\mu_{2}, \Sigma_{2})$ with a high-dimensional appearance $a_{1}$.

We synthesize the output image of the subject $I_{2} \in \mathbb{R}^{H_{O} \times W_{O} \times 3}$ using the primitive image $J_{2} \in \mathbb{R}^{H_{R} \times W_{R} \times A}$, which is fed to an encoder-decoder network.
Following StyleGAN2~\cite{karras2020analyzing} style mixing and design principles, we use a U-Net encoder-decoder~\cite{ronneberger2015u}.
Where the output resolution is higher than the rendered resolution, the input is upsampled.
To recover high-frequency details, we incorporate the appearance $a_{1}$ as styles at each stage,
\begin{align}
    \label{eq:Enhancing} J_{2}, a_{1} \rightarrow I_{2}
\end{align}
From the input image $I_{1}^{*}$, we only estimate information about the foreground subject.
As it is impossible to accurately infer a novel view of a static background captured by a static camera, we infer a constant background around the subject, and use a segmentation mask to discard the background information from the groundtruth image $I_{2}^{*}$.

\subsection{Losses}
\label{sec:MethodologyLosses}

\subsubsection{Image Reconstruction}
We assess the performance of novel view synthesis by measuring the average pixel-to-pixel distance between the image generated by the model $I_{2}$ and the target image $I_{2}^{*}$ in the pixel space,
\begin{align}
\label{eq:LossPixel}
    \mathcal{L}_{I} = \mathbb{E}_{I_{1}^{*}, I_{2}^{*}}\left[{||I_{2}^{*} - I_{2}||}_{1}^{1}\right]
\end{align}
Which we complement with either the standard perceptual loss~\cite{johnson2016perceptual}, $\mathcal{L}_{\phi_{\text{VGG}}}$ with a pretrained VGG network~\cite{simonyan2014very}, or the fine-tuned LPIPS loss~\cite{zhang2018unreasonable}, $\mathcal{L}_{\phi_{\text{LPIPS}}}$, to enable the model to synthesize images containing high-frequency detail.

\subsubsection{Adversarial Framework}
To further enhance the realism of the synthesized images, we fine-tune our novel synthesis model in an adversarial framework,
\begin{align}
    \mathcal{L}_{A} = \mathbb{E}_{I_{2}^{*}}\left[\log(D(I_{2}^{*}))\right] + \mathbb{E}_{I_{1}^{*}}\left[\log(1 - D(I_{2}))\right]
\end{align}

\subsubsection{Pose Estimation}
We require supervision to ensure that the locations of body parts correspond to prespecified keypoints and convey the same semantic meaning,
\begin{align}
    \label{eq:LossPose}
    \mathcal{L}_{P} = \mathbb{E}_{p_{1}^{*}, \bar{P}_{1}^{*}}\left[{c^{*} \cdot ||\bar{P}_{1}^{*} - \bar{P}_{1}||}_{2}^{2}\right]
\end{align}
Where $c^{*} \in \mathbb{R}^{N}$ denotes the confidence values of the points.

\subsubsection{Appearance Regularization}
Appearance vectors are an unsupervised intermediate representation. We regularize the squared norm of the appearance vectors,
\begin{align}
    \label{eq:RegularisationAppearance}
    R_{a} = {||a_{1}||}_{2}^{2}
\end{align}

\subsubsection{Final Objective}
We obtain our final objective,
\begin{align}
    \label{eq:LossTotal}
    \min_{M} \max_{D} \lambda_{I} \mathcal{L}_{I} + \lambda_{\phi} \mathcal{L}_{\phi} + \lambda_{A} \mathcal{L}_{A} + \lambda_{P} \mathcal{L}_{P} + \lambda_{a} R_{a}
\end{align}
With, $\lambda_{I} = \lambda_{\phi} = 10$, $\lambda_{A} = \lambda_{P} = 1$, and $\lambda_{a} = 10^{-3}$.

%% file: figures/infer_real.tex
\begin{figure}[tbp]
    \centering
    \includegraphics[width=\linewidth]{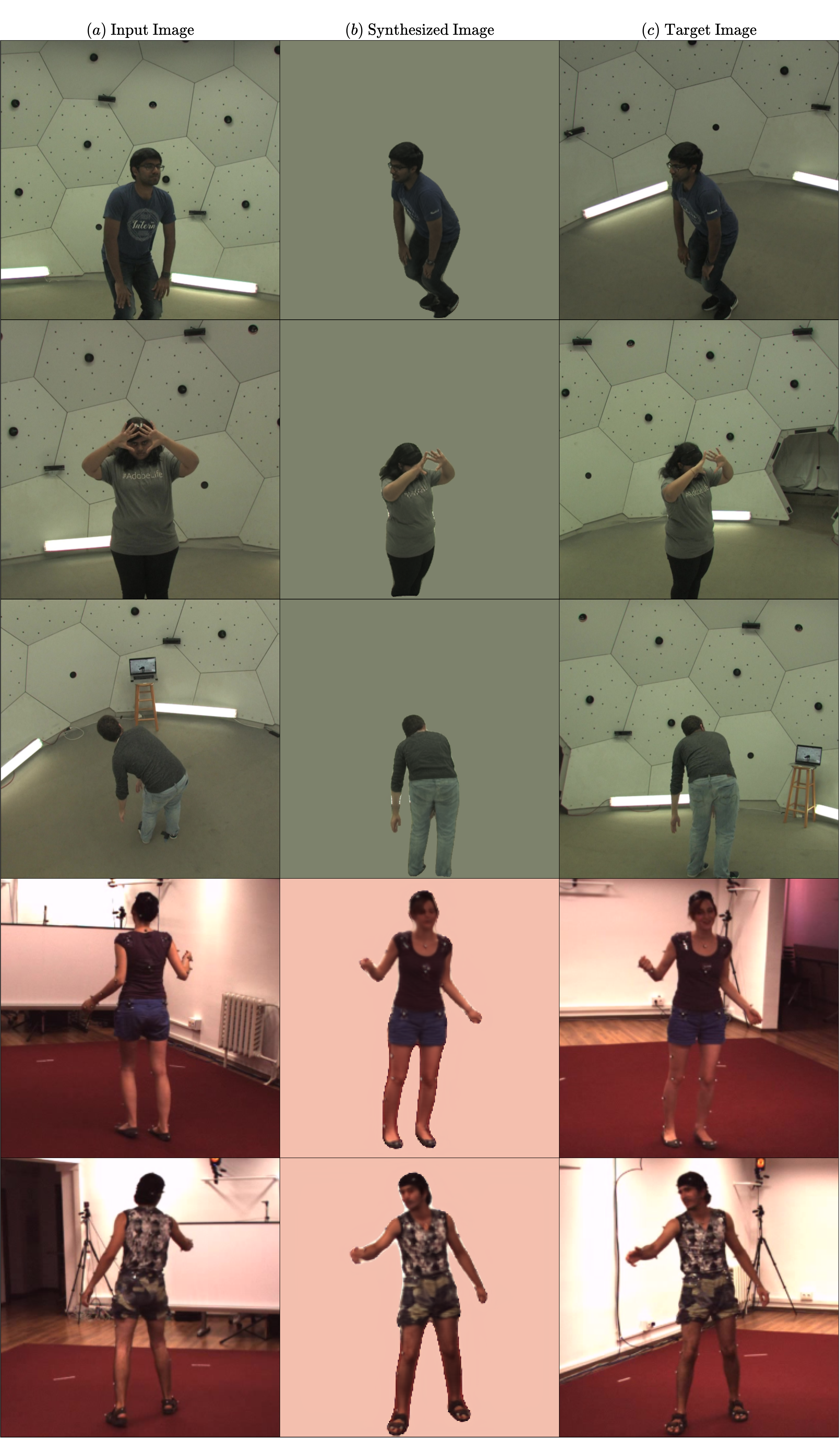}
    \caption{View synthesis of known subjects in previously unseen poses on Panoptic Studio \textit{(top)} and Human3.6M \textit{(bottom)}. From the input image $(a)$, we extract pose and appearance before rendering the primitives in the novel viewpoint and synthesizing the image $(b)$ which resembles target image $(c)$.}
    \label{fig:NovelViewSynthesis}
\end{figure}

%% file: sections/experiments.tex
\section{Experiments}
\label{sec:Experiments}

Our framework is implemented in PyTorch and trained end-to-end, with the exception of the 2D detector, which is OpenPose~\cite{cao2017realtime}.
Our models are trained with a batch size of $32$, using AdamW~\cite{loshchilov2017decoupled}, with a learning rate of $2 \cdot 10^{-3}$ and a weight decay of $10^{-1}$.
The appearance vectors are 16 dimensional. 
For the differentiable renderer, we set $\alpha = 2.5 \cdot 10^{-2}$, $\beta = 2$ and $b = 0_{A}$.
The segmentation masks used for cropping the background out of the groundtruth image are obtained with SOLOv2~\cite{wang2020solov2}.
Images are loosely cropped around the subject to a resolution of ${1080}^{2}$ for Panoptic Studio and cropped given a bounding box of the subject for Human3.6M.
Input images are resized to a resolution of ${256}^{2}$, high-dimensional latent images are rendered at a resolution of ${256}^{2}$, and output images are resized to either ${256}^{2}$ or ${512}^{2}$.

\subsection*{Datasets}
\label{sec:Datasets}

\subsubsection*{Panoptic Studio}
 Joo \etal~\cite{joo2017panoptic} provides marker-less multi-view sequences captured in a studio.
There are over 70 sequences captured from multiple cameras, including 31 HD cameras at 30 Hz.
We restrict ourselves to single person sequences, and use only images recorded by high-definition cameras.
Panoptic Studio presents greater variability in subject's clothing, morphology and ethnicity, with the most significant detail being that it was captured in a marker-less fashion, compared to datasets such as Human3.6M~\cite{ionescu2013human3}.
Its high variability in camera poses is beneficial both for novel view synthesis and when demonstrating the 3D understanding and robustness of the model. 

We infer 2D poses by running OpenPose over every view of a sequence, and reconstruct 3D pose following the approach detailed by Faugeras~\cite{faugeras1993three}, which computes a closed-form initialization followed by an iterative refinement.
We remove consistently unreliable 2D estimates and obtain a $117$-point body model. 

Data is partitioned at the subject level into training, validation, and test sets, with each subject in only one of the sets.
Approximately $80\%$ of the frames are used for training, $10\%$ for validation and $10\%$ for testing. This corresponds to 29 subjects for training, 4 for validation, and 4 for testing.
We require pairs of images for training, therefore we have $|\mathcal{E}| = |\mathcal{F}| \times {|\mathcal{V}|}_{2}$ possible pairs, with $\mathcal{F}$ and $\mathcal{V}$, referring to the sets of available frames and views respectively.
This gives us a total of $81.6$M pairs of images for training, $11.7$M pairs for validation and $12.7$M for testing.

\subsubsection*{Human3.6M}
Human3.6M~\cite{ionescu2013human3} is one of the largest publicly available datasets for 3D human pose estimation.
It contains sequences of 11  actors performing 15 different scenarios recorded with 4 high-definition cameras at 50 Hz.
Mo-Cap is used to provide  accurate ground truth.
We use the provided $17$-joint skeletal model.
Following previous work, data is partitioned at the subject level into training (S1, S5, S6, S7 and S8) and validation on two subjects (S9 and S11).

\subsection{Ablation Study}
\label{sec:ExperimentsAblationStudy}

\begin{table*}[tbp]
    \centering
    \caption{Comparison of the reconstruction quality depending on the losses.}
    \begin{tabular}{@{}lccccccccc@{}}
    \toprule
     & \multicolumn{3}{c}{Panoptic@256} & \multicolumn{3}{c}{Panoptic@512} & \multicolumn{3}{c}{Human3.6M@256} \\
    \cmidrule(lr){2-4} \cmidrule(lr){5-7} \cmidrule(lr){8-10}
     & LPIPS $\downarrow$  & PNSR $\uparrow$ & SSIM $\uparrow$  & LPIPS $\downarrow$  & PNSR $\uparrow$ & SSIM $\uparrow$ & LPIPS $\downarrow$  & PNSR $\uparrow$ & SSIM $\uparrow$ \\
    \midrule
    $\mathcal{L}_{I}$                                           & 0.0512            & \textbf{32.34}    & \textbf{0.9553}   & 0.0578            & \textbf{31.67}    & \textbf{0.9572}   & 0.1168            & \textbf{21.47}    & \textbf{0.8738} \\
    + $\mathcal{L}_{\phi_{\text{VGG}}}$                         & 0.0363            & 32.15             & 0.9540            & 0.0403            & 31.49             & 0.9554            & 0.0905            & 21.39             & 0.8720 \\
    + $\mathcal{L}_{\phi_{\text{LPIPS}}}$                       & 0.0291            & 31.96             & 0.9512            & 0.0323            & 31.13             & 0.9521            & \textbf{0.0792}   & 21.17             & 0.8669 \\
    + $\mathcal{L}_{\phi_{\text{LPIPS}}}$ + $\mathcal{L}_{A}$   & \textbf{0.0284}   & 31.91             & 0.9519            & \textbf{0.0318}   & 31.01             & 0.9509            & -                 & -                 & -      \\
    \bottomrule
    \end{tabular}
    \label{tab:AblationStudy}
\end{table*}

\input{figures/ablation_study}

We evaluate the contribution of each loss for the view synthesis task, reporting the LPIPS, PSNR and SSIM.
For each sequence, we sample two subsequences representing $10\%$ of the whole sequence, for validation and testing, and use the remaining frames for training.
We select models by looking at the validation error, and report the results on the test set.
As expected, \cref{tab:AblationStudy} shows that, the models trained with the LPIPS loss achieve a better LPIPS score, while the models trained with the pixel and VGG losses have lower PSNR and SSIM error.
However, when we look closely at the qualitative examples, we can see on \cref{fig:AblationStudy} that the models trained with the LPIPS loss contain much more high-frequency detail, and that the adversarial loss enables to reach finer levels of details in images.
We notice a sharp gap in performance between the models trained on Panoptic Studio and Human3.6M, which we attribute to the granularity of skeletal model, on Panoptic Studio, the skeletal model provides detailed information about the hands and face, while on Human3.6M, face and hands are represented by a single point.
More examples are given in the supplementary material.

\subsection{Motion Transfer}
\label{sec:ExperimentsMotionTransfer}

\input{figures/motion_transfer}

To demonstrate the versatility of our approach, we apply it to whole body motion transfer.
\Cref{fig:MotionTransfer} shows how our approach can be used for motion and view transfer from one individual to another.
Given an unseen person $(a)$ from the test partition, we  estimate their pose from a new viewpoint, and extract the appearance of an individual $(b)$ whose style was learned during training.
Since our framework naturally disentangles pose and appearance, without further training, we can combine them and render the image from a novel viewpoint and obtain $(c)$.
We provide $(d)$ as a visual comparison to show that our network is able to extract and render 3D information faithfully.
Despite small errors in the pose, caused by ambiguities in the pose image $(a)$, we reconstruct a convincing approximation of a different person in the same pose.
As such, this work represents an initial step towards full-actor synthesis from arbitrary input videos.

\subsection{Synthesizing Images from Unseen Viewpoints}

\input{figures/infer_virtual}

As our model is trained on multiple views, it can generate realistic looking images of a subject in unseen poses from virtual viewpoints, \eg where cameras do not actually exist.
As seen in \cref{fig:VirtualViews}, we can create virtual cameras travelling on a spherical orbit around the subject. More videos examples are given in the supplementary material.

\subsection{Learning Novel View Synthesis of Unknown Subjects from a Monocular Sequence}

\input{figures/appearance_learning}

While our model generalizes well to the pose estimation task, the number of subjects (29 and 5) in both datasets are insufficient for the model to generalize over the space of all human appearances.
However, using a monocular sequence of an unseen subject, we can quickly retrain both the appearance and image synthesis modules to the new individual.
Our model is able to produce novel views of an unseen individual from a single camera.
As we see in \cref{fig:AppearanceLearning} appearance fine-tuning shows a clear visual improvement, however it is more effective on Human3.6M than Panoptic Studio. This is a consequence of subjects facing in one direction only for single sequences in Panoptic Studio. As such, novel view synthesis requires estimating the appearance of completely unseen parts.

%% file: figures/ablation_study.tex
\begin{figure}[tbp]
    \centering
    \includegraphics[width=\linewidth]{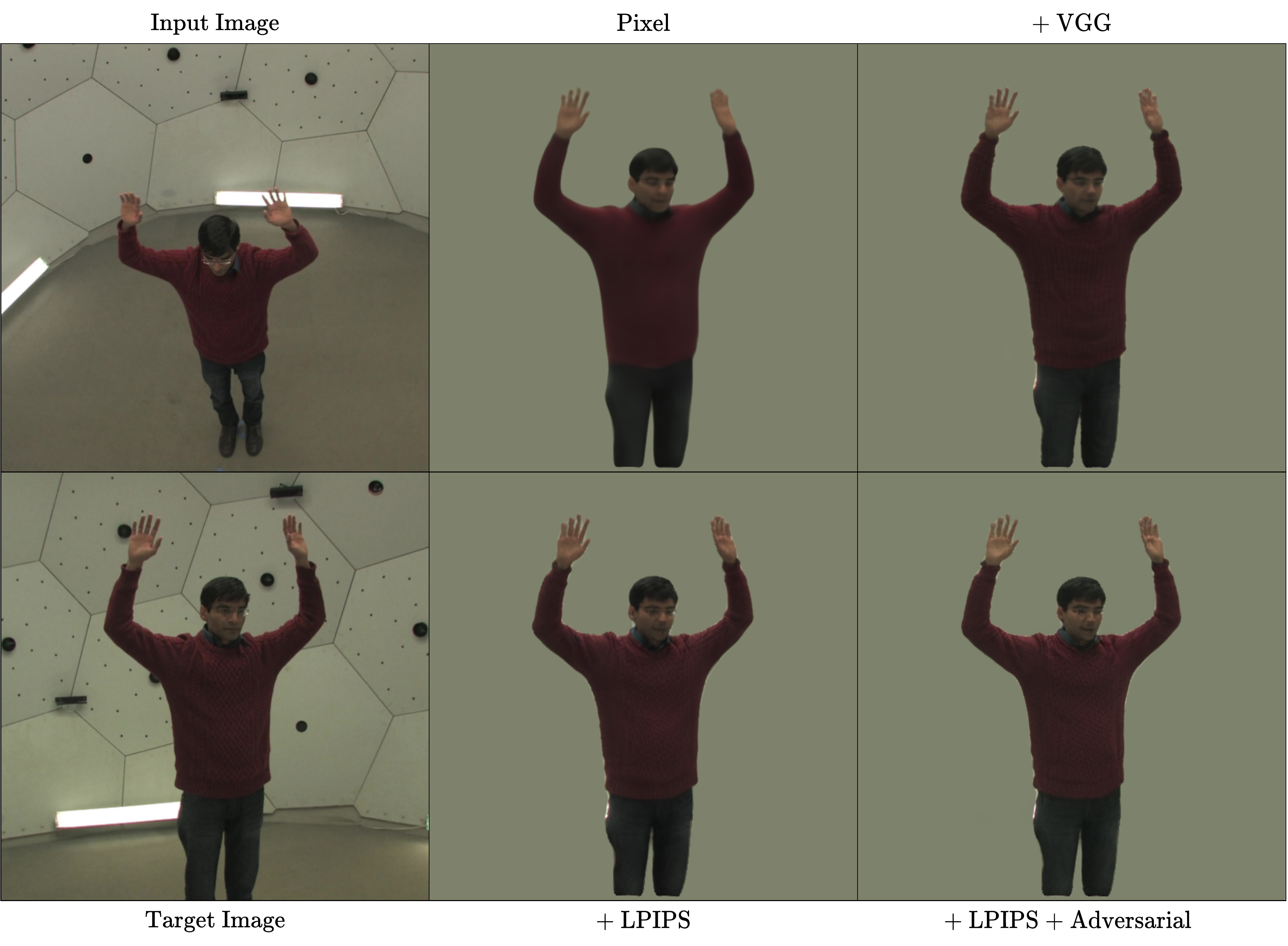}
    \caption{Effects of the losses on the view synthesis of known subjects in previously unseen poses. There is a clear improvement in reconstruction quality when adding perceptual losses, and the highest-frequency detail appears with the adversarial training.}
    \label{fig:AblationStudy}
\end{figure}

%% file: figures/motion_transfer.tex
\begin{figure}[tbp]
    \centering
    \includegraphics[width=\linewidth]{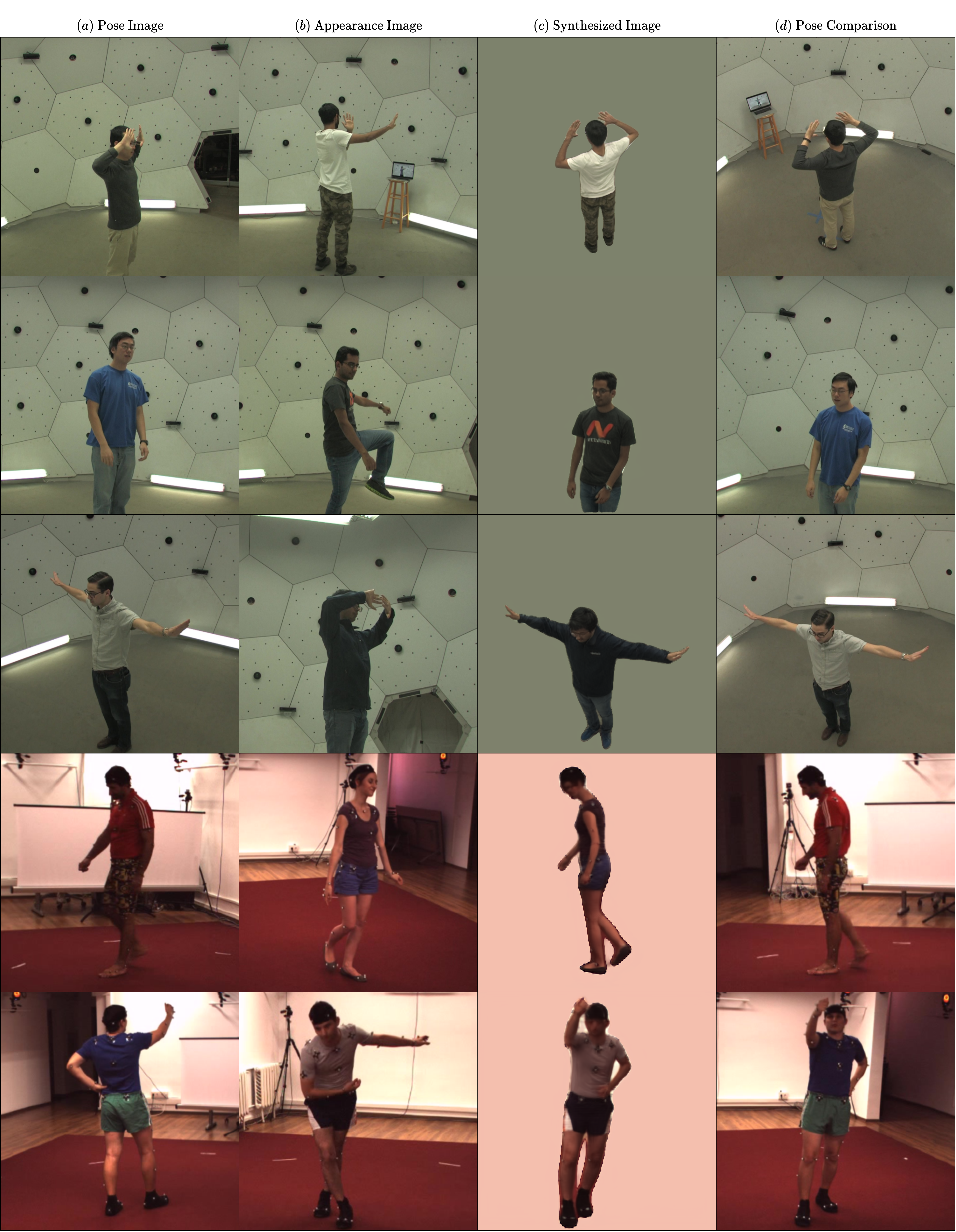}
    \caption{Motion transfer on Panoptic Studio and Human3.6M. We extract pose from an previously unseen subject $(a)$ and appearance from a known subject $(b)$, and synthesize the combination into a novel view $(c)$, while $(d)$ shows the similarity in terms of pose.}
    \label{fig:MotionTransfer}
\end{figure}

%% file: figures/infer_virtual.tex
\begin{figure}[tbp]
    \centering
    \includegraphics[width=\linewidth]{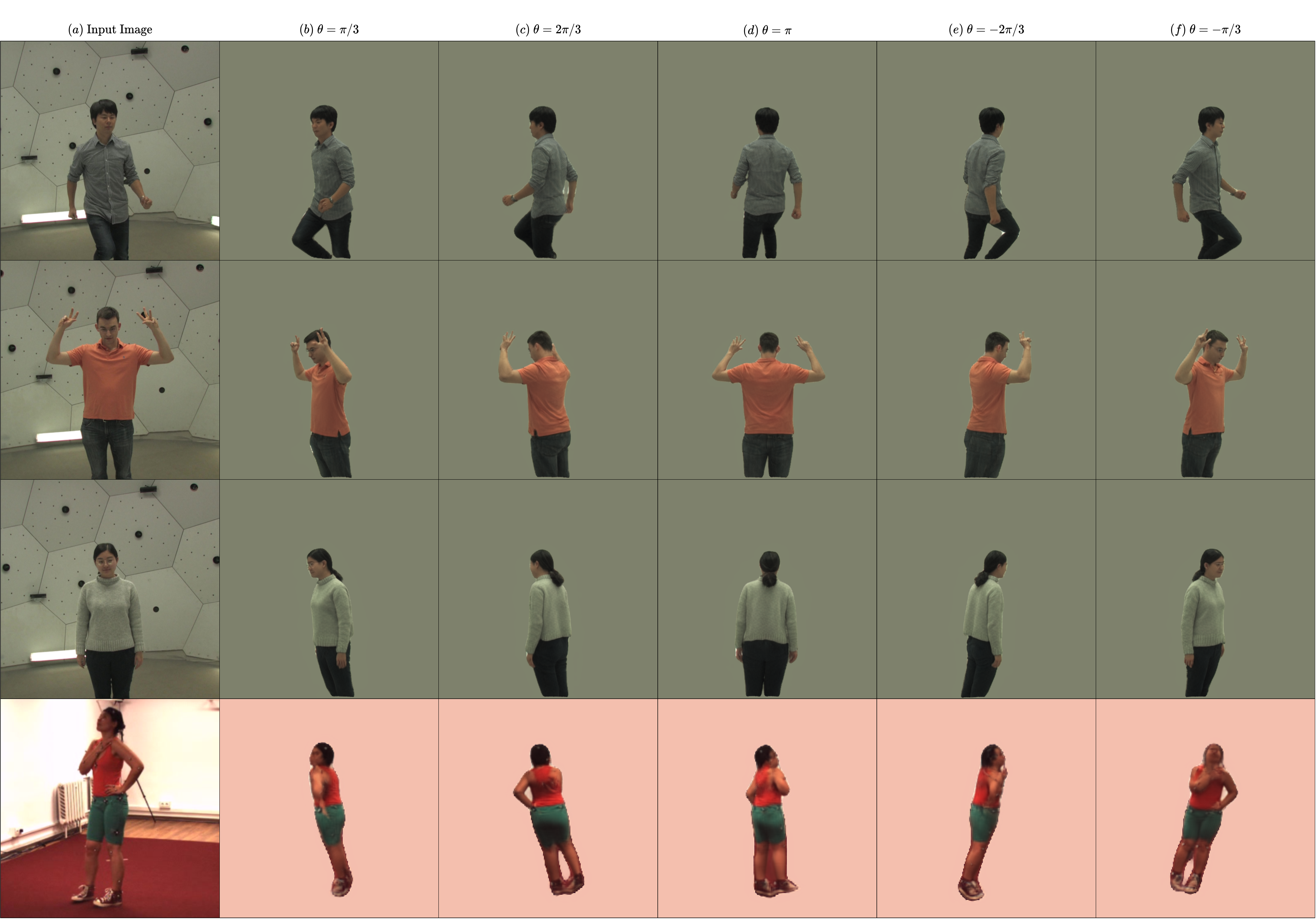}
    \caption{Given an input image $(a)$ of a known subject in an unknown pose, we synthesize novel views from non-existing viewpoints on a spherical orbit $(b - f)$ on Panoptic Studio and Human3.6M.}
    \label{fig:VirtualViews}
\end{figure}

%% file: figures/appearance_learning.tex
\begin{figure}[tbp]
    \centering
    \includegraphics[width=\linewidth]{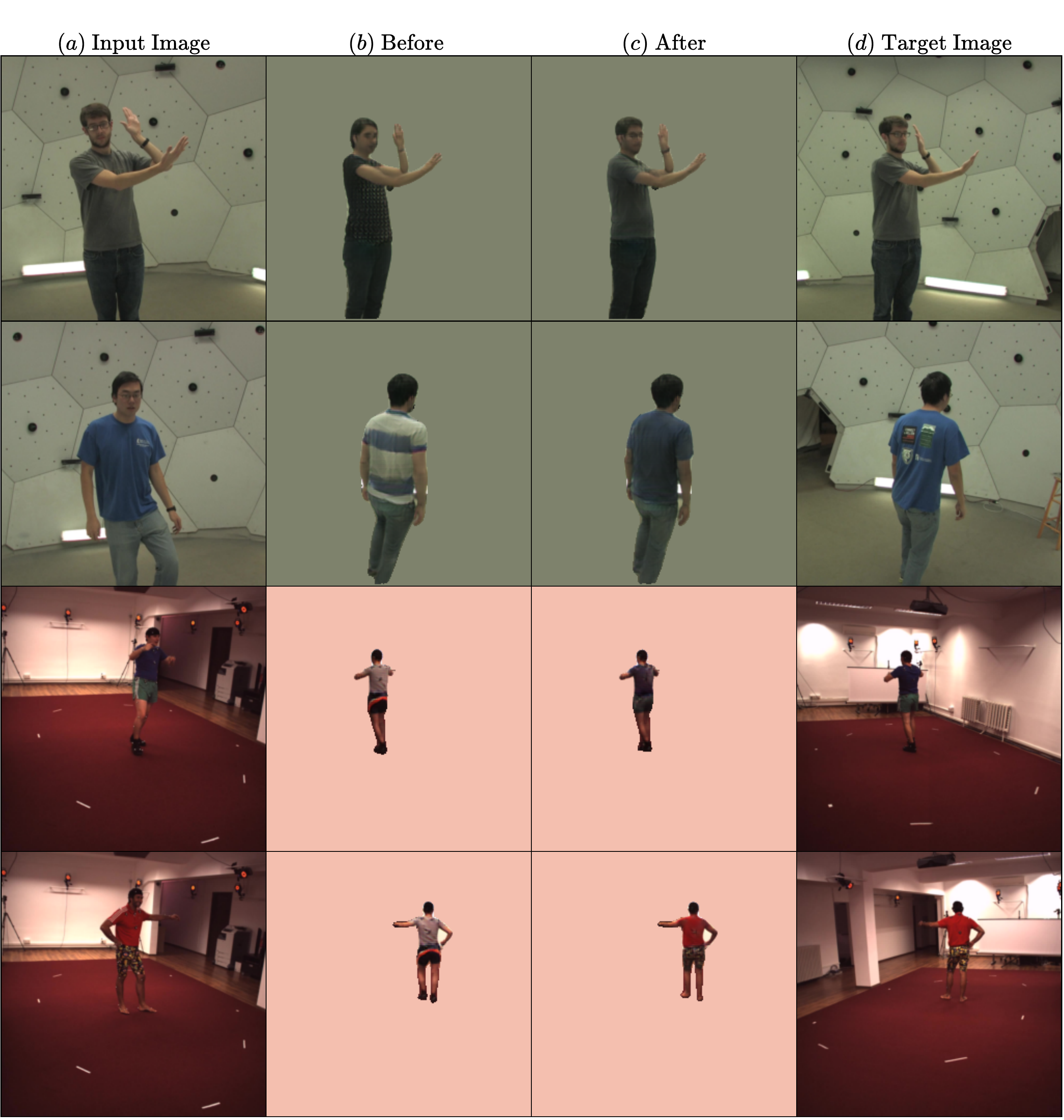}
    \caption{We fine-tune the appearance and image synthesis module on a monocular sequence of a previously unseen subject. Given an input image $(a)$ appearance from an unknown subject, we synthesize novel views before $(b)$ and after fine-tuning $(c)$.}
    \label{fig:AppearanceLearning}
\end{figure}

%% file: sections/conclusion.tex
\section{Conclusion}
\label{sec:Conclusion}
We have presented a novel 3D renderer highly suited for human reconstruction and synthesis.
By design, our formulation encodes a semantically and physically meaningful latent 3D space of parts while our novel feature rendering approach translates these parts into 2D images giving rise to a robust and easy to optimize representation.
A encoder-decoder architecture allows us to transfer style and move from feature images back into the image space.
We illustrated the versatility of our approach on multiple tasks: semi-supervised learning; novel view synthesis; and style and motion transfer, allowing us to puppet one person's movement from any viewpoint.

Although we focus on human reconstruction, we believe our approach of rendering a small number of tractable and semantically meaningful primitives as feature images to be useful in a wider scope.
Potential applications include semantic mapping, reconstruction and dynamic estimation of the poses of articulated objects and animals, and novel view synthesis of rigid objects.
To this end, we will release the complete source code.

%% file: sections/acknowledgements.tex
\section{Acknowledgements}
This work received funding from the SNSF Sinergia project ‘SMILE II’ (CRSII5 193686), the European Union’s Horizon2020 research and innovation programme under grant agreement no. 101016982 ‘EASIER’ and the EPSRC project ‘ExTOL’ (EP/R03298X/1). This work reflects only the authors view and the Commission is not responsible for any use that may be made of the information it contains.

%% file: sections/appendix.tex
\section{Finding the Depth of the Root Joint}
\label{appendix:Depth}
We have $p^{*} \in \mathbb{R}^{J \times 3}$, the 2D pose in ray coordinates, and $\bar{P}^{*} \in \mathbb{R}^{(J - 1) \times 3}$, the 3D pose in the camera view, centered around the root joint. \\
We want to find $Z_{\text{root}}^{*} \in \mathbb{R}^{+}$ to obtain $P \in \mathbb{R}^{J \times 3}$, the 3D pose in the camera coordinate frame,
\begin{align}
    P = [P_{\text{root}} | P_{\text{root}} + \bar{P}]
\end{align}
With,
\begin{align}
    P_{\text{root}} = Z_{\text{root}}^{*} \cdot p_{\text{root}}^{*}
\end{align}
Where,
\begin{align}
    Z_{\text{root}}^{*} = \argmin_{Z_{\text{root}}} L \implies \nabla_{Z_{\text{root}}^{*}}{L} = 0
\end{align}
We develop,
\begin{align}
    L
    =
    \frac{1}{J - 1} \sum_{j}{d_{j}}
    \implies
    \nabla_{Z_{1}}{L}
    =
    \frac{1}{J - 1} \sum_{j}{\nabla_{Z_{1}}{d_{j}}}
\end{align}
And,
\begin{align}
    d_{j}
    =
    {||p_{j} - p_{j}^{*}||}_{2}^{2}
    \implies
    \nabla_{p_{j}}{d_{j}}
    =
    2 \cdot (p_{j} - p_{j}^{*})
\end{align}
And,
\begin{align}
    p_{j}
    =
    \begin{pmatrix}
    x_{j} \\
    y_{j} \\
    1
    \end{pmatrix}
    =
    \begin{pmatrix}
    X_{j} / Z_{j} \\
    Y_{j} / Z_{j} \\
    1
    \end{pmatrix}
    \implies
    \nabla_{P_{j}}{p_{j}}
    =
    \begin{pmatrix}
    \frac{1}{Z_{j}} & 0 & -\frac{X_{j}}{Z_{j}^{2}} \\
    0 & \frac{1}{Z_{j}} & -\frac{Y_{j}}{Z_{j}^{2}} \\
    0 & 0 & 0
    \end{pmatrix}
\end{align}
And,
\begin{align}
    P_{j}
    =
    \begin{pmatrix}
    X_{j} \\
    Y_{j} \\
    Z_{j}
    \end{pmatrix}
    =
    \begin{pmatrix}
    X_{1} + \bar{X}_{j}^{*} \\
    Y_{1} + \bar{Y}_{j}^{*} \\
    Z_{1} + \bar{Z}_{j}^{*}
    \end{pmatrix}
    \implies
    \nabla_{P_{1}}{P_{j}}
    =
    \begin{pmatrix}
    1 & 0 & 0 \\
    0 & 1 & 0 \\
    0 & 0 & 1
    \end{pmatrix}
\end{align}
And,
\begin{align}
    P_{1}
    =
    Z_{1} \cdot p_{1}^{*}
    =
    \begin{pmatrix}
    Z_{1} \cdot x_{1}^{*} \\
    Z_{1} \cdot y_{1}^{*} \\
    Z_{1}
    \end{pmatrix}
    \implies
    \nabla_{Z_{1}}{P_{1}}
    =
    \begin{pmatrix}
        x_{1}^{*} \\
        y_{1}^{*} \\
        1
    \end{pmatrix}
\end{align}
By the chain rule,
\begin{align}
    \nabla_{Z_{1}}{L}
    &=
    \frac{1}{J - 1} \sum_{j = 2}^{J}
    {
    \nabla_{p_{j}}{d_{j}}
    \times
    \nabla_{P_{j}}{p_{j}}
    \times
    \nabla_{P_{1}}{P_{j}}
    \times
    \nabla_{Z_{1}}{P_{1}}
    } \\
    &=
    \frac{1}{J - 1} \sum_{j = 2}^{J}
    {
    \frac{2}{Z_{j}} [(x_{j} - x_{j}^{*})(x_{1}^{*} - x_{j}) + (y_{j} - y_{j}^{*})(y_{1}^{*} - y_{j})]
    } \\
    &= \frac{1}{J - 1} \sum_{j = 2}^{J}
    {
    \frac{2}{Z_{1} + \bar{Z}_{j}^{*}}
    \left[\left(\frac{Z_{1} \cdot x_{1}^{*} + \bar{X}_{j}^{*}}{Z_{1} + \bar{Z}_{j}^{*}} - x_{j}^{*}\right) \cdot \left(x_{1}^{*} - \frac{Z_{1} \cdot x_{1}^{*} + \bar{X}_{j}^{*}}{Z_{1} + \bar{Z}_{j}^{*}}\right)
    + \left(\frac{Z_{1} \cdot y_{1}^{*} + \bar{Y}_{j}^{*}}{Z_{1} + \bar{Z}_{j}^{*}} - y_{j}^{*}\right) \cdot \left(y_{1}^{*} - \frac{Z_{1} \cdot y_{1}^{*} + \bar{Y}_{j}^{*}}{Z_{1} + \bar{Z}_{j}^{*}}\right)\right]
    }
\end{align}
Using a symbolic solver, we obtain,
\begin{align}
    Z_{1}^{*}
    =
    \frac{1}{J - 1} \sum_{j = 2}^{J}
    {
    \frac
    {
    \bar{X}_{j}^{*2} + \bar{Y}_{j}^{*2} +
    [
    (x_{j}^{*} \cdot x_{1}^{*} + y_{j}^{*} \cdot y_{1}^{*}) \cdot \bar{Z}_{j}^{*}
    - (x_{j}^{*} + x_{1}^{*}) \cdot \bar{X}_{j}^{*}
    - (y_{j}^{*} + y_{1}^{*}) \cdot \bar{Y}_{j}^{*}
    ]
    \cdot \bar{Z}_{j}^{*}
    }
    {
    (x_{j}^{*} - x_{1}^{*}) \cdot (\bar{X}_{j}^{*} - x_{1}^{*} \cdot \bar{Z}_{j}^{*})
    + (y_{j}^{*} - y_{1}^{*}) \cdot (\bar{Y}_{j}^{*} - y_{1}^{*} \cdot \bar{Z}_{j}^{*})
    }
    }
\end{align}

\section{Finding the Rotation Between Two Vectors}
\label{appendix:Rotation}
We want to find the rotation matrix between two non-zero vectors, $x, y \in \mathbb{R}^{n} \setminus \{0\}$.
\\
We normalise $x$ to a unit vector $u$,
\begin{align}
    u = \frac{x}{||x||}
\end{align}
We compute the normalised vector rejection $v$ of $y$ on $u$,
\begin{align}
    v = \frac{y - (u^{\intercal} y) u}{||y - (u^{\intercal} y) u||}
\end{align}
We compute the cosinus and sinus of the angle $\theta$ between $x$ and $y$,
\begin{align}
    \cos(\theta)    &=  \frac{x^{\intercal} y}{||x|| \cdot ||y||} \\
    \sin(\theta)    &= \sqrt{1 - \cos^{2}(\theta)}
\end{align}
We compute the projection $Q$ onto the complemented space generated by $x$ and $y$,
\begin{align}
    Q = I - u u^{\intercal} + v v^{\intercal}
\end{align}
Finally, we compute the rotation $R$,
\begin{align}
    R = Q +
    {\begin{bmatrix}
        u \\
        v
    \end{bmatrix}}^{\intercal}
    {\begin{pmatrix}
        \cos(\theta)    &   -\sin(\theta)\\
        \sin(\theta)    &   \cos(\theta)
    \end{pmatrix}}
    {\begin{bmatrix}
        u \\
        v
    \end{bmatrix}}
\end{align}

\section{Differentiable Rendering of Diffuse Primitives}
\label{appendix:DifferentiableRendering}
Modelling a scene as a collection of diffuse primitives, we render a high-dimensional latent image,
\begin{align}
    J &= \mathcal{R}_{\alpha, \beta}(\mu, \Sigma, a, b, K, D) \in \mathbb{R}^{H \times W \times A}
\end{align}
Where,
\begin{itemize}
    \item $\mathcal{R}$ is the rendering function;
    \item $\alpha > 0$ is a coefficient scaling the magnitude of the shapes of the primitives;
    \item $\beta > 1$ is a background blending coefficient;
    \item $\mu = \{\mu_{k} \in \mathbb{R}^{3} | k \in [1..M]\}$, where $\mu_{k}$ refers to the location of the $k^\text{th}$ primitive in camera coordinates;
    \item $\Sigma = \{\Sigma_{k} \in \mathbb{R}^{3 \times 3} | k \in [1..M]\}$, where $\Sigma_{k}$ refers to the ellipsoidal shape of the $k^\text{th}$ primitive, given by its positive definite matrix;
    \item $a = \{a_{k} \in \mathbb{R}^{A} | k \in [1..M]\}$, where $a_{k}$ describes the appearance of the $k^\text{th}$ primitive;
    \item $b \in \mathbb{R}^{A}$, describes the appearance of the background;
    \item $K \in \mathbb{R}^{3 \times 3}$ and $D \in \mathbb{R}^{K}$ refers to the intrinsic parameters and distortion coefficients.
\end{itemize}
To simplify notation, we retain the following letters for subscripts: $i \in [1..H]$ refers to the height of the image, $j \in [1..W]$ refers to the width of the image, and $k \in [1..M]$ refers to each of the $M$ primitives.

We define the rays $r_{i j}$ as unit vectors originating from the pinhole, distorted by the lens, and passing through every pixels of the image,
\begin{align}
    r_{i j} &= \frac{u(K^{-1} \times p_{i j}, D)}{{||u(K^{-1} \times p_{i j}, D)||}_{2}}
\end{align}
Where, $u$ is a fast fixed-point iterative method finding an approximate solution for un-distorting the rays, and $p_{i j} = \begin{pmatrix} j & i & 1 \end{pmatrix}^{\intercal}$ is a pixel on the image plane.

Let $F_{i j k}$ be the density diffused onto a ray $r_{i j}$ by a single primitive $(\mu_{k}, \Sigma_{k})$,
\begin{align}
    F_{i j k} &= \int_{0}^{+\infty}{e^{-\Delta^{2}(z \cdot r_{i j}, \mu_{k}, \alpha \cdot \Sigma_{k})} dz}
\end{align}
See \cref{appendix:Rendering_Integral} for the analytical solution.

For each ray $r_{i j}$, we define a smooth rasterisation coefficient $\lambda_{i j k}$ for each primitive $(\mu_{k}, \Sigma_{k})$.
In a nutshell, this coefficient `smoothly' favours a primitive, and discounts the others, based on their proximity to the ray $r_{i j}$.
See \cref{appendix:Smooth_Rasterization} for details.

The background is treated as the $(M + 1)^{\text{th}}$ primitive, with unique properties.
Its density $F_{i j M + 1}$ and smooth rasterisation coefficient $\lambda_{i j M + 1}$ are detailed in \cref{appendix:Background_Primitive}.

We derive the weights $\omega_{i j k}$ quantifying the influence of each primitive $(\mu_{k}, \Sigma_{k})$ (including the background) onto each ray $r_{i j}$, $\forall k \in [1..{M + 1}]$,
\begin{align}
    \omega_{i j k} &= \frac{\lambda_{i j k} \cdot F_{i j k}}{\sum\limits_{l=1}^{M + 1}{\lambda_{i j l} \cdot F_{i j l}}}
\end{align}
Finally, we render the image by combining the weights with their respective appearance,
\begin{align}
    J_{i j} &= \sum\limits_{k=1}^{M + 1}{\omega_{i j k} \cdot a_{k}}
\end{align}

\subsection{Rendering Integral}
\label{appendix:Rendering_Integral}
We want to measure the density $F_{i j k}$ diffused by a single primitive $(\mu_{k}, \Sigma_{k})$ along a ray $r_{i j}$, by calculating the integral of the diffusion along a ray,
\begin{align}
    F_{i j k} &= \int_{0}^{+\infty}{e^{-\Delta^{2}(z \cdot r_{i j}, \mu_{k}, \alpha \cdot \Sigma_{k})} dz}
\end{align}
To simplify the notation further, we remove the subscript notations. \\
Let,
\begin{align}
    F &= \int_{0}^{+\infty}{e^{-\Delta^{2}(z \cdot r, \mu, \alpha \cdot \Sigma)} dz}
\end{align}
With $\Delta^{2}$, the squared Mahalanobis distance,
\begin{align}
    \Delta^{2}(u, v, A) &= \sbf{\left(u - v\right)}{A}{\left(u - v\right)}
\end{align}
We define $\Sigma'$ to simplify future calculations,
\begin{align}
    \Sigma' &= \alpha \cdot \Sigma
\end{align}
We expand and factorise the quadratic form to isolate $z$,
\begin{align}
    \Delta^{2}(z \cdot r, \mu, \alpha \cdot \Sigma)
    &=
    \Delta^{2}(z \cdot r, \mu, \Sigma') \\
    &=
    \sbf{\left(z \cdot r - \mu\right)}{\Sigma'}{\left(z \cdot r - \mu\right)} \\
    &=
    z^2 \sbf{r}{\Sigma'}{r} - 2 z \sbf{r}{\Sigma'}{\mu} + \sbf{\mu}{\Sigma'}{\mu} \\
    &=
    \sbf{r}{\Sigma'}{r} \left[z^2 - 2 z\frac{\sbf{r}{\Sigma'}{\mu}}{\sbf{r}{\Sigma'}{r}} \right] + \sbf{\mu}{\Sigma'}{\mu} \\
    &=
    \sbf{r}{\Sigma'}{r} \left[z^2 - 2 z\frac{\sbf{r}{\Sigma'}{\mu}}{\sbf{r}{\Sigma'}{r}} + {\left(\frac{\sbf{r}{\Sigma'}{\mu}}{\sbf{r}{\Sigma'}{r}} \right)}^2 - {\left(\frac{\sbf{r}{\Sigma'}{\mu}}{\sbf{r}{\Sigma'}{r}} \right)}^2\right] + \sbf{\mu}{\Sigma'}{\mu} \\
    &=
    \sbf{r}{\Sigma'}{r} \left[ {\left(z - \frac{\sbf{r}{\Sigma'}{\mu}}{\sbf{r}{\Sigma'}{r}} \right)}^2 - {\left(\frac{\sbf{r}{\Sigma'}{\mu}}{\sbf{r}{\Sigma'}{r}} \right)}^2 \right] + \sbf{\mu}{\Sigma'}{\mu} \\
    &=
    \sbf{r}{\Sigma'}{r} {\left(z - \frac{\sbf{r}{\Sigma'}{\mu}}{\sbf{r}{\Sigma'}{r}} \right)}^2 - \frac{{\left(\sbf{r}{\Sigma'}{\mu}\right)}^2}{\sbf{r}{\Sigma'}{r}} + \sbf{\mu}{\Sigma'}{\mu}
\end{align}
Therefore,
\begin{align}
    F(r, \mu, \alpha \cdot \Sigma)
    &=
    \int_{0}^{+\infty}{e^{-\Delta^{2}(z \cdot r, \mu, \alpha \cdot \Sigma)} dz} \\
    &=
    \int_{0}^{+\infty}{e^{-\Delta^{2}(z \cdot r, \mu, \Sigma')} dz} \\
    &=
    \int_{0}^{+\infty}{e^{- \left[\sbf{r}{\Sigma'}{r} {\left(z - \frac{\sbf{r}{\Sigma'}{\mu}}{\sbf{r}{\Sigma'}{r}} \right)}^2 - \frac{{\left(\sbf{r}{\Sigma'}{\mu}\right)}^2}{\sbf{r}{\Sigma'}{r}} + \sbf{\mu}{\Sigma'}{\mu}\right]} dz} \\
    &=
    e^{- \left[- \frac{{\left(\sbf{r}{\Sigma'}{\mu}\right)}^2}{\sbf{r}{\Sigma'}{r}} + \sbf{\mu}{\Sigma'}{\mu}\right]} \int_{0}^{+\infty}{e^{- \sbf{r}{\Sigma'}{r} {\left(z - \frac{\sbf{r}{\Sigma'}{\mu}}{\sbf{r}{\Sigma'}{r}} \right)}^2} dz}
\end{align}
Let,
\begin{align}
    u &= \sqrt{\sbf{r}{\Sigma'}{r}} {\left(z - \frac{\sbf{r}{\Sigma'}{\mu}}{\sbf{r}{\Sigma'}{r}} \right)}
\end{align}
Therefore,
\begin{align}
    du &= \sqrt{\sbf{r}{\Sigma'}{r}} dz
\end{align}
\begin{align}
    u(0) &= \frac{\sbf{r}{\Sigma'}{\mu}}{\sqrt{\sbf{r}{\Sigma'}{r}}} \\
    \lim_{z \to +\infty} u(z) &= +\infty
\end{align}
By substitution,
\begin{align}
    F(r, \mu, \Sigma')
    &=
    e^{- \left[- \frac{{\left(\sbf{r}{\Sigma'}{\mu}\right)}^2}{\sbf{r}{\Sigma'}{r}} + \sbf{\mu}{\Sigma'}{\mu}\right]} \frac{1}{\sqrt{\sbf{r}{\Sigma'}{r}}} \int_{u(0)}^{+\infty}{e^{- u^{2}} du}
\end{align}
However, we know that,
\begin{align}
    \erfc(x) &= \frac{2}{\sqrt{\pi}} \int_{x}^{+\infty}{e^{- u^{2}} du}
\end{align}
Therefore,
\begin{align}
    F(r, \mu, \Sigma')
    &=
    e^{- \left[\sbf{\mu}{\Sigma'}{\mu} - \frac{{\left(\sbf{r}{\Sigma'}{\mu}\right)}^2}{\sbf{r}{\Sigma'}{r}}\right]} \frac{1}{\sqrt{\sbf{r}{\Sigma'}{r}}} \frac{\sqrt{\pi}}{2} \erfc(u(0)) \\
    &=
    \frac{\sqrt{\pi}}{2} \frac{1}{\sqrt{\sbf{r}{\Sigma'}{r}}} \erfc\left(\frac{\sbf{r}{\Sigma'}{\mu}}{\sqrt{\sbf{r}{\Sigma'}{r}}}\right)  e^{-\left[\sbf{\mu}{\Sigma'}{\mu} - \frac{{\left(\sbf{r}{\Sigma'}{\mu}\right)}^2}{\sbf{r}{\Sigma'}{r}}\right]}
\end{align}
Finally,
\begin{align}
    F(r, \mu, \alpha \cdot \Sigma)
    &=
    \frac{\sqrt{\alpha \pi}}{2 \sqrt{\sbf{r}{\Sigma}{r}}} \erfc\left( \frac{\sbf{r}{\Sigma}{\mu}}{\sqrt{\alpha} \sqrt{\sbf{r}{\Sigma}{r}}}\right)  e^{-\frac{1}{\alpha} \left[\sbf{\mu}{\Sigma}{\mu} - \frac{{\left(\sbf{r}{\Sigma}{\mu}\right)}^2}{\sbf{r}{\Sigma}{r}}\right]}
\end{align}

\subsection{Smooth Rasterization}
\label{appendix:Smooth_Rasterization}

Objects closer to the camera should occlude those farther from it. \\
For each ray $r_{i j}$, we define a smooth rasterisation coefficient $\lambda_{i j k}$ for each primitive $(\mu_{k}, \Sigma_{k})$,
\begin{align}
    \lambda_{i j k} &=  \frac{1}{1 + {(z_{i j k}^{*})}^{4}}
\end{align}
Where,
\begin{align}
    z_{i j k}^{*} &= \argmax_{z} {e^{-\Delta^{2}(z \cdot r_{i j}, \mu_{k}, \alpha \cdot \Sigma_{k})}}
\end{align}
The optimal depth $z_{i j k}^{*}$, is attained when the squared Mahalanobis distance $\Delta^{2}$ between the point on the ray $z \cdot r_{i j}$ and the primitive $(\mu_{k}, \Sigma_{k})$ is minimal, which in turns maximises the Gaussian density function.

Again, to simplify the notation further, we remove the subscript notations. \\
Let,
\begin{align}
    z^{*} &= \argmax_{z} {e^{-\Delta^{2}(z \cdot r, \mu, \alpha \cdot \Sigma)}}
\end{align}
With,
\begin{align}
    \Delta^{2}(z \cdot r, \mu, \alpha \cdot \Sigma)
    &=
    \Delta^{2}(z \cdot r, \mu, \Sigma') \\
    &=
    \sbf{\left(z \cdot r - \mu\right)}{\Sigma'}{\left(z \cdot r - \mu\right)} \\
    &=
    \sbf{r}{\Sigma'}{r} {\left(z - \frac{\sbf{r}{\Sigma'}{\mu}}{\sbf{r}{\Sigma'}{r}} \right)}^2 - \frac{{\left(\sbf{r}{\Sigma'}{\mu}\right)}^2}{\sbf{r}{\Sigma'}{r}} + \sbf{\mu}{\Sigma'}{\mu}
\end{align}
We solve,
\begin{align}
    \frac{\partial e^{-\Delta^{2}}}{\partial z} &= 0
\end{align}
By the chain rule,
\begin{align}
    \frac{\partial e^{-\Delta^{2}}}{\partial z} &= \frac{\partial e^{-\Delta^{2}}}{\partial \Delta^{2}} \frac{\partial \Delta^{2}}{\partial z}
\end{align}
We have,
\begin{align}
    \frac{\partial \Delta^{2}}{\partial z} &= 2 \sbf{r}{\Sigma'}{r} {\left(z - \frac{\sbf{r}{\Sigma'}{\mu}}{\sbf{r}{\Sigma'}{r}} \right)}
\end{align}
And,
\begin{align}
    \frac{\partial e^{-\Delta^{2}}}{\partial \Delta^{2}} &= - e^{-\Delta^{2}} < 0
\end{align}
Therefore,
\begin{align}
    \frac{\partial e^{-\Delta^{2}}}{\partial z} = 0 \iff \frac{\partial \Delta^{2}}{\partial z}   =  0
\end{align}
Which implies,
\begin{align}
    z^{*} &= \frac{\sbf{r}{\Sigma'}{\mu}}{\sbf{r}{\Sigma'}{r}} = \frac{\sbf{r}{\left( \alpha \cdot \Sigma \right)}{\mu}}{\sbf{r}{\left( \alpha \cdot \Sigma \right)}{r}}
\end{align}
Therefore,
\begin{align}
    z^{*} &= \frac{\sbf{r}{\Sigma}{\mu}}{\sbf{r}{\Sigma}{r}}
\end{align}
Finally, we apply the density function to this distance to obtain the smooth rasterization coefficient, in order to model the order in which primitives should appear,
\begin{align}
    \lambda &= \frac{1}{1 + {(z^{*})}^{4}}
\end{align}

\subsection{The Background Primitive}
\label{appendix:Background_Primitive}
We consider the background as an additional primitive, \emph{e.g.} the $(M+1)^\text{th}$, with special properties.
First, it is colinear with every ray and located after the furthest primitive,
\begin{align}
    \mu_{i j {M + 1}} &= z^{*}_{M + 1} \cdot r_{i j}
\end{align}
With,
\begin{align}
    z^{*}_{M + 1} &= \beta \cdot \max_{i j k}{z^{*}_{i j k}}
\end{align}
Where $\beta$ determines how further away the background is assumed to be, we set $\beta = 2$. \\
Second, its shape is a constant, given by the identity matrix,
\begin{align}
    \Sigma_{M + 1} &= \alpha \cdot I
\end{align}
Therefore its density is given by,
\begin{align}
    F_{i j {M + 1}}
    &=
    \int_{0}^{+\infty}{e^{-\Delta^{2}(r_{i j}, \mu_{i j {M + 1}}, \Sigma_{M + 1})} dz}
\end{align}
Which simplifies to,
\begin{align}
    F_{i j {M + 1}}
    &=
    \frac{\sqrt{\alpha \pi}}{2} \erfc\left(\frac{z^{*}_{M + 1}}{\sqrt{\alpha}}\right)
\end{align}
As we can see, its density is not tied to the rays, but to the depth of the primitives only. \\
Third, its rasterization coefficient is given by,
\begin{align}
    \lambda_{M + 1} &= \frac{1}{1 + {(z^{*}_{M + 1})}^{4}}
\end{align}
Last, it has a given constant appearance,
\begin{align}
    a_{M + 1} &= b
\end{align}